\documentclass[10pt,twocolumn,letterpaper]{article}
\usepackage{amsfonts,amssymb}

\usepackage{iccv}
\usepackage{times}
\usepackage{epsfig}
\usepackage{graphicx}
\usepackage{amsmath}
\usepackage{amssymb}
\usepackage{algorithm}
\usepackage{algorithmic}
\usepackage{xcolor}
\usepackage{url}
\usepackage{footmisc}
\usepackage{booktabs}  
\usepackage{multirow}
\usepackage{float}
\usepackage{graphicx}
\usepackage{bbm}
\usepackage{CJK}
\usepackage{threeparttable}
\usepackage{graphics}
\usepackage{epsfig}

\usepackage[breaklinks=true,bookmarks=false]{hyperref}

\iccvfinalcopy 

\def\iccvPaperID{} 

\ificcvfinal\fi
\input{Definitions}
\graphicspath{{figures/}}

\begin{document}

\title{Generation For Adaption: A GAN-Based Approach for Unsupervised Domain Adaption with 3D Point Cloud Data }

\author{ Junxuan Huang

 \qquad Junsong Yuan  \qquad Chunming Qiao \\
State University of New York at Buffalo\\
{\tt\small \{junxuanh, jsyuan ,qiao\}@buffalo.edu }}
\maketitle
\ificcvfinal\thispagestyle{empty}\fi

\begin{abstract}
Recent deep networks have achieved good performance on a variety of 3d points classification tasks. However, these models often face challenges in “wild tasks” where there are considerable differences between the labeled training/source data collected by one Lidar and unseen test/target data collected by a different Lidar. Unsupervised domain adaptation (UDA) seeks to overcome such a problem without target domain labels. Instead of aligning features between source data and target data, we propose a method that uses a Generative Adversarial Network (GAN) to generate synthetic data from the source domain so that the output is close to the target domain. Experiments show that our approach performs better than other state-of-the-art UDA methods in three popular 3D object/scene datasets (i.e., ModelNet, ShapeNet and ScanNet) for cross-domain 3D object classification.
\end{abstract}

\section{Introduction}
\begin{figure}
	\centering
	
	\includegraphics[width=0.95\columnwidth]{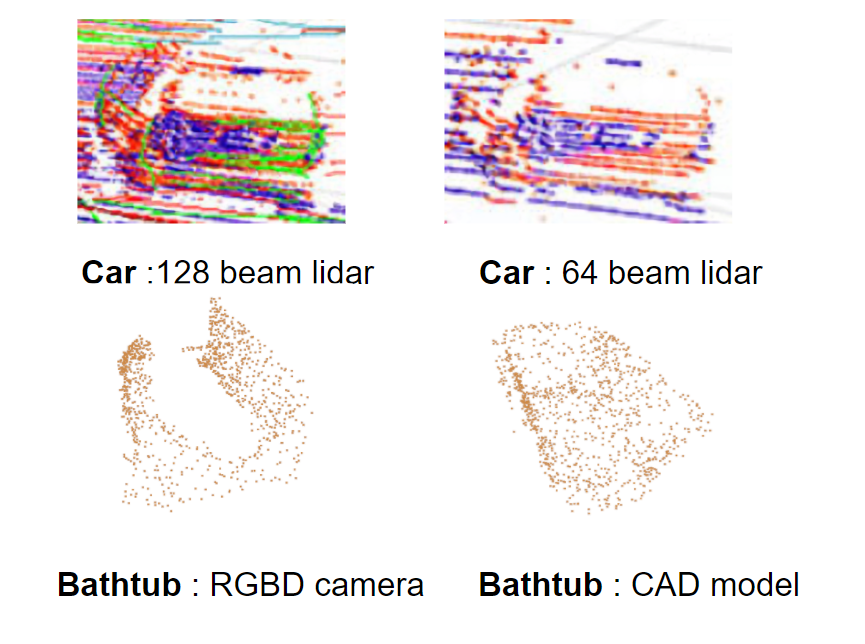}
	
	\caption{Point cloud generate from different source or sensors may have different geometric feature. Even the same car scanned by two different lidar have different points density. }
	\label{fig:domain gap}
	\vspace{-0.3cm}
	
\end{figure}

Deep learning with 3D point cloud data has achieved significant outcomes in different tasks. Classification is the most fundamental task \cite{qi2017pointnet} and plays a central role in a number of
modern applications, e.g., robots, self-driving cars or virtual reality. Despite their impressive success, a Deep Neural Network (DNN) requires a large amount of labeled point cloud data for training. Point clouds data can be captured with different sensors (e.g. 64 beam lidar, 128 beam lidar and RGB-D camera) which produces different 3D sampling patterns. In addition, in real-world where objects are scanned from LiDAR, it also happens that some parts are lost or occluded (e.g.chairs lose legs) while some datasets are generated form 3D polygonal models with a smooth and uniform surface. As Fig.~\ref{fig:domain gap} shows, apparently, those datasets have different geometric feature and data distribuations. Which may leads to a performance drop when we adapting a classification model. This issue is known as \textit{domain gap}.

To solve the performance dropping issue, labeling every 3D object in the unseen dataset is the most straightforward solution. But different with 2D object dataset. During labeling, people need to rotate several times and look through different angles to identify an object. It is time-consuming and expensive. The objective of domain adaptation (DA) is to leverage rich annotations in a source domain to achieve a good performance in a target domain having few annotations. Some of the existing DA methods have focused on mapping features into a shared subspace or minimizing instance-level distances such as MMD \cite{borgwardt2006integrating}, CORAL \cite{sun2016deep}. Other adversarial-training DA methods, like DANN \cite{ganin2014unsupervised}, ADDA \cite{tzeng2017adversarial}, MCD \cite{saito2018maximum} have tried to use adversarial-training to select domain invariant features during training so that the trained model could perform better in the target domain. The main idea of domain adversarial training is that if the feature representation is domain invariant, a classifier trained on the source domain’s features will operate on the target domain as well. However, the assumption about the presence of domain invariant features may not always be true.
As an alternative, PointDan \cite{pointdan} designed a Self-Adaptive Node Construction for aligning 3D local features with points cloud data. A recent work \cite{yi2020complete} built a DA method that upsampled the source domain data into a canonical domain and trained a semantic segmentation network over the canonical domain.

 \begin{figure}
	\centering
	
	\includegraphics[width=0.95\columnwidth]{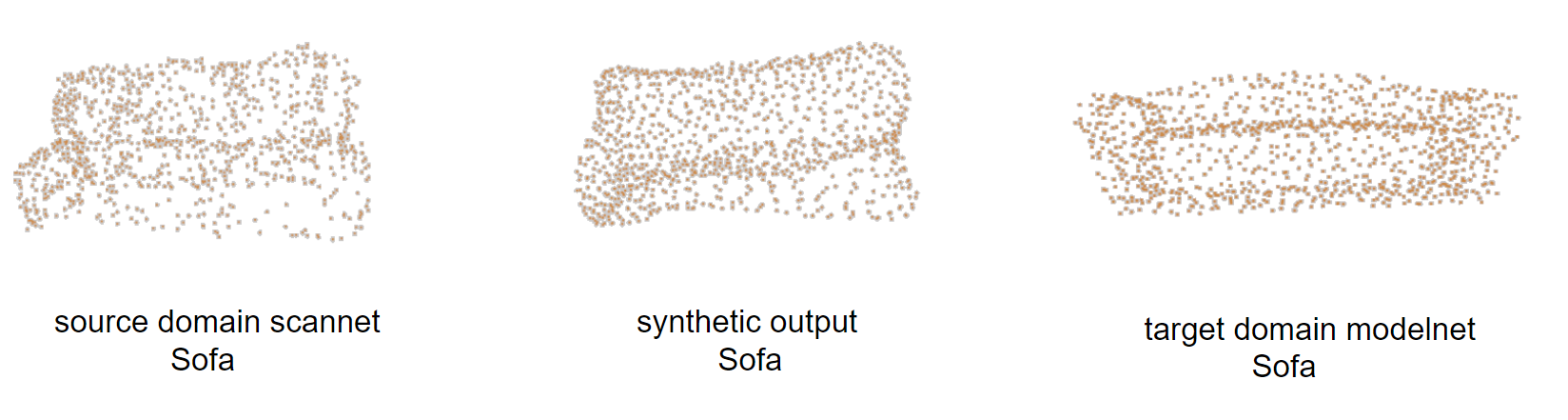}
	
	\caption{The distribution of points cloud in target object (right) looks much uniform than source object (left), and our synthetic output keep the shape of the source domain object, but the distribution of points are more similar to the target domain object }
	\label{fig:first_example}
	\vspace{-0.3cm}
	
\end{figure}

In this paper, we tackle the problem of unsupervised domain adaptation (UDA), where the target domain is entirely unlabelled. Our approach chooses to generate synthetic data from a source domain according to the target domain data pattern. Unlike other adversarial-training DA methods, we fully utilize the advantage of adversarial-training by generating synthetic data while keeping the label of the source domain. As Fig.~\ref{fig:first_example} shows, our model generates a synthetic sofa with a very similar shape to the sofa from the source dataset and its point distribution looks much closer to the target dataset object compared with the sofa from the source dataset. 
Our model architecture consists of four parts: generator,latent reconstruct module, discriminator, and feature encoder/decoder. We have two inputs for the generator, in addition to the encoded feature of input points cloud, we also have a latent code $\mathit{z}$ sampled from standard Gaussian distribution as a condition input. Because of that, we can generate multi possible outputs from a single input, and this can improve the quality of output object from the generator as shown in \cite{wu_2020_ECCV}. Latent reconstruct module recovered the latent code $\mathit{z}$ during training for calculating latent reconstruct loss. For the encoder, we use Point transformer structure \cite{guo2020pct} and decoder uses a 3-layer MLP (multilayer perceptron). Similar to D2GAN \cite{nguyen2017dual}, we have an additional discriminator, a classifier to restrict the output from generator. Extensive experiments that involve the use of different pairs of a dataset to perform cross-domain 3D object classification tasks show that our approach's predicting accuracy is 2.6\% higher than PointDan on average, 2.8\% higher than MCD, and 4.6\% higher than ADDA.

We summarize our contributions as follows: 
\begin{itemize}
  \item [(1)] 
 A new UDA model for 3D object classification which different from other methods in this task. We generate synthetic data for training classification models instead of searching common feature space or aligning features. Based on that, we can visualize the effect of domain adaption and indicate which part of the object is transferred    
  \item [(2)]
  A  novel design that adding a latent space reconstruct module and using an additional discriminator to restrict the generator's output which further improves the adaptation.
  \item [(3)]
  We perform extensive experimental tests on various 3D classification datasets and show that the proposed approach can achieve competitive results compared to the state-of-the-art methods.
\end{itemize}

\section{Related Work}
\subsection{3D Vision Understanding}
3D vision has various data representations: multi-view, voxel grid, 3D mesh and point cloud data. Various Deep networks have been designed to deal with the above different formats of 3D data~\cite{su2015multi,maturana2015voxnet,you2018pvnet, feng2019meshnet}. Point cloud can be directly obtained by LiDAR sensors, and contain a lot of 3D spatial information. PointNet \cite{qi2017pointnet}, which takes advantage of a symmetric function to process the unordered point sets in 3D, is the first deep neural networks to directly deal with point clouds. Later research~\cite{qi2017pointnet++} proposed to stack PointNets hierarchically to model neighborhood information and increase model capacity. Several subsequent works considered how to perform convolution operations like 2D images on point clouds. 

One main approach is to convert a point cloud into a regular voxel array to allow convolution operations. Tchapmi et al.~\cite{Tchapmi2017segcloud} proposed SEGCloud for pointwise segmentation. It maps convolution features of 3D voxels to point clouds using trilinear interpolation and keeps global consistency through fully connected conditional random fields. Atzmon et al~\cite{Atzmon2018point} presented the PCNN framework with extension and restriction operators to map between point-based representation and voxel-based representation by performing volumetric convolution on voxels to extract point features. MCCNN by Hermosilla et al.~\cite{hermosilla2018monte} allows non-uniformly sampled point clouds by treating convolution as a Monte Carlo integration problem. Similarly, in PointConv proposed by Wu et al.~\cite{Wu2019pointconv}, 3D convolution was performed through Monte Carlo estimation and importance sampling. Liu et al.~\cite{liu2019pointvoxel} took advantages of volumetric 3D convolution and point-based method by proposing point-voxel convolution.

\subsection{ Unsupervised Domain Adaptation (UDA)}
 Early work on UDA mainly performed reweighing~\cite{sugiyama2008direct,zhang2013domain} or resampling~\cite{gong2013connecting} of the source-domain examples to match the target distribution. Besides, there has been a prolific body of works on learning domain-invariant representations. For instance, Fernando et al~\cite{fernando2013unsupervised} tried to use subspace alignment and Gong et al~\cite{gong2017geodesic} performed interpolation. Adversarial training~\cite{ganin2016domain,tzeng2017adversarial,bousmalis2017unsupervised,shrivastava2017learning,hoffman2017cycada} tried to use a discriminator to select domain-invariant features. Other methods such as Correlation Alignment (CORAL) \cite{sun2016deep}, Maximum Mean Discrepancy (MMD)~\cite{borgwardt2006integrating,long2013transfer}, or Geodesic distance~\cite{gopalan2011domain} tried to minimize the intra-class distance in a subspace simultaneously. However, there have been only a few works on domain adaptation for 3D point clouds. Among them. Rist et al.~\cite{rist2019cross} proposed that dense 3D voxels are preferable to point clouds for sensor-invariant processing of LiDAR point clouds. Salah et al.~\cite{saleh2019domain} proposed a CycleGAN approach to the adaptation of 2D bird's eye view images of LiDAR between synthetic and real domains. Wu et al.~\cite{wu2019squeezesegv2} used geodesic correlation alignment between real and synthetic data. Qin et al.~\cite{pointdan} designed a Self-Adaptive Node Construction for aligning 3D local features with points cloud data.

 \begin{figure}
	\centering
	
	\includegraphics[width=0.95\columnwidth]{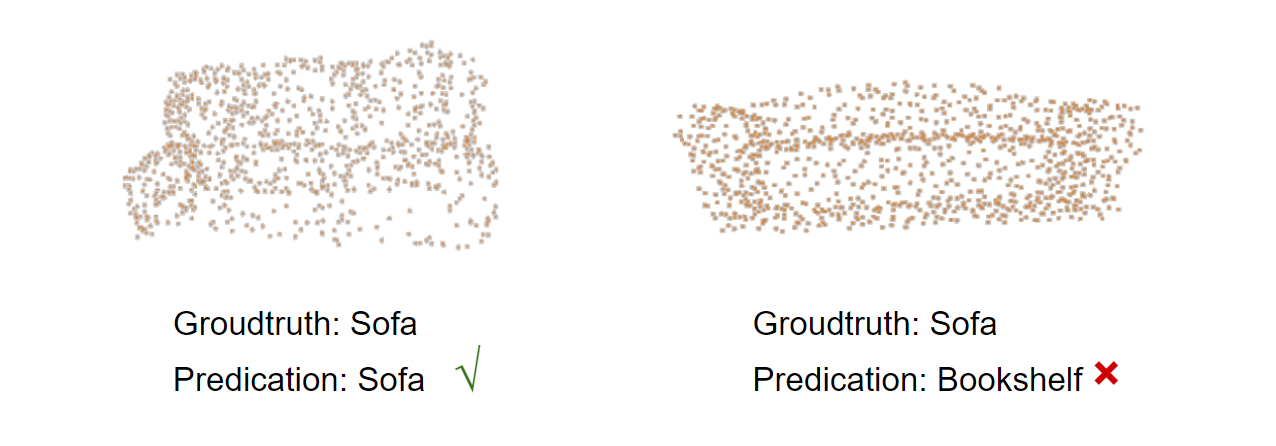}
	
	\caption{A 3D classification model trained on source dataset Scannet (left) make a wrong prediction on object from target dataset Modelnet (right) due to domain gap between two datasets }
	\label{fig:wrong_case}
	\vspace{-0.3cm}
	
\end{figure}

\section{Proposed GAN-based DA method}
As Fig.~\ref{fig:wrong_case} shows, when we directly use a classifier trained on the source dataset over the target dataset, some objects may be misclassified due to the domain gap. A sofa from Modelnet may be classified as a bookshelf if we use a classifier trained on Scannet without any domain adaption methods. To overcome the domain gap between different point cloud datasets, we propose a generative model, which takes point cloud from the source dataset and target dataset as input first and then tries to maintain the shape of an object while keeping the label information when generating the synthetic object according to the target domain dataset's distribution. This adaption process's objective is that classifier could learn the distribution of point cloud in the target domain and perform better in the test of the target dataset.  

\begin{figure*}
	\centering
	\includegraphics[width=1.6\columnwidth]{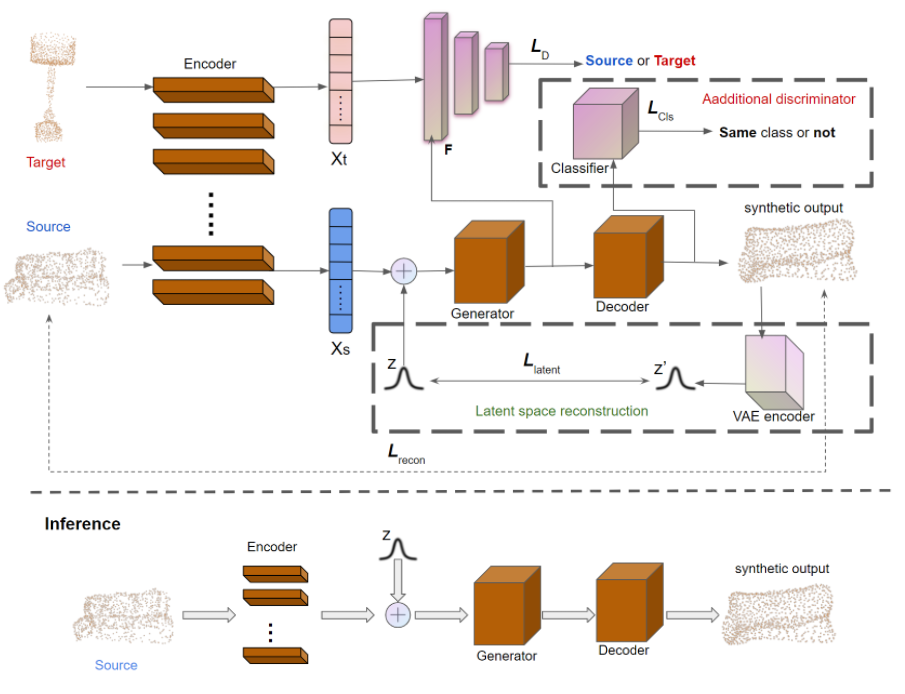}
	
	\caption{Illustration of training and inference  .}
	\label{fig:structure}
	\vspace{-0.3cm}
\end{figure*}

As Fig.~\ref{fig:structure} shows in the training source domain object and target domain object will go through a shared encoder. Like most GANs, the encoded features will be sent to a discriminator and, the discriminator tries to distinguish features from source domain or target domain. Following the idea in \cite{wu_2020_ECCV} of adding multimodal information to the model, we also have Gaussian samples $\mathit{z}$ for latent condition input to the generator. To force the generator to use the Gaussian samples $\mathit{z}$, we introduce a VAE encoder to recover $\mathit{z}$ from the synthetic output. In addition, inspired by \cite{liu2019generative}, in order to enhance the quality of output object from $\Gb$, we have an additional discriminator, a classifier $\Cb$ in training the model.Below, we explain our model in more details.

\subsection{Definition and Notations}\label{sec:pointcloud}
We consider an unsupervised domainadaptation (UDA) setting, 
A point cloud from a source domain is represented as $(\Xb^s\triangleq\{\xb_i^s\}_{i = 1...N}, y^s)$, where $\xb_i^s \in \mathbb{R}^3$ is a 3D point and $N$ is the number of points in the point cloud; $y \in \{1, 2,...,k\}$ is the ground-truth label, where $k$ is the number of classes.
In UDA, we have access to a set of labeled LiDAR point clouds, from the source domain and a set of unlabeled LiDAR point clouds 
a set of unlabeled LiDAR point clouds $(\Xb^t\triangleq\{\xb_i^t\}_{i = 1...N})$ in a target domain.
 It is assumed that two domains are sampled from the distributions $P_s(\Xb^s)$ and $P_t(\Xb^t)$ while the  distribution  $P_s \neq P_t$.
We use classification as the task of domain adaption and denote the classification network as $\Cb_\thetab(\cdot) \triangleq \{C_{\thetab, j}| j = 1...k\}$, whose input is a point cloud $\Xb$ and output is a probability vector $\Cb_\thetab(\Xb)$.
our approach tend to generate synthetic object $\Xb'^s$ from $\Xb^s$ and for the generative generative adversarial network we have encoder $\Eb$, decoder $\Db$, generator $\Gb$ and discriminator $\Fb$. The process is denoted as $\Xb'=D(G(E(\Xb)))$

\subsection{Learn mapping of point cloud to latent space}
\begin{figure}
    \centering
	\includegraphics[width=0.45\columnwidth]{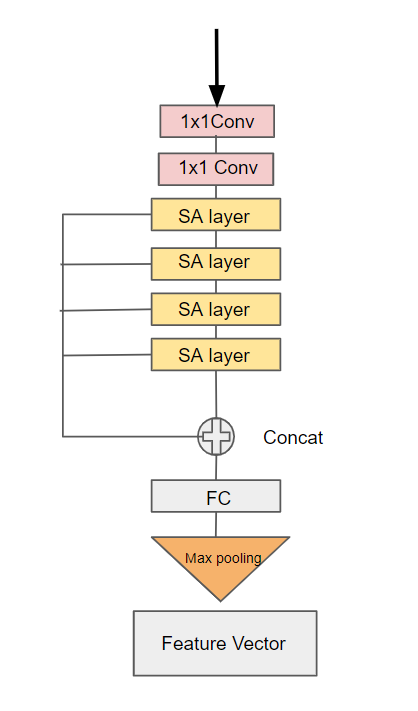}
	
	\caption{The structure of encoder}
	\label{encoder}
	\vspace{-0.3cm}
\end{figure}

We obtain feature vector $\mathbb{X}_s$ of input point cloud by training an autoencoder. As Fig.~\ref{encoder} shows the encoder $\Eb$  consists of 4 self-attention layer \cite{guo2020pct}. The object $\Xb^s$ from the source dataset is encoded as feature vector $\mathbb{X}_s$ and decoder $\Db$ reconstructs object $\tilde\Xb$ from the latent feature vector $\mathbb{X}_s$. The encoder and decoder are trained with the reconstruction loss, and we choose Earth Mover’s Distance (EMD) to measure the distance between reconstructed object $\tilde\Xb$ and input object $\Xb^s$.
\begin{equation}
\mathcal{L}^{\textrm{recon}} =d^{\text{EMD}}( \Xb^s,  \Db (\Eb (\Xb^s))), 
\end{equation}
As for the object $\Xb^t$ from the target dataset, instead of training another autoencoder for the target domain, we directly feed $\Xb^t$  to $\Eb$ because  \cite{chen2020} showed that doing this would achieve a better performance in subsequent adversarial training.
\subsection{Latent space reconstruction}

Like most of the generative adversarial networks, we set a min-max game between generator and discriminator. The generator is trained to fool the discriminator so that the discriminator fails to tell if the latent vector comes from the source domain $\Xb^s$ or target domain $\Xb^t$. 
Use Earth Mover's Distance(EMD) to restrict the synthetic outputs can make it close to the input's shape, but we do not want the synthetic object to having the exact shape of input. Because we are building a synthetic dataset which means the variety is also significant. So we bring a random sampled variable \textbf{z} into our model and
a Variational Autoencoder(VAE) is trained to encode synthetic objects to recover latent input vector, encouraging the use of conditional mode input \textbf{z}.

Formally, the latent representation of the source domain input  $\xb_s=E_{\text{AE}}(\Xb^s)$, along with random sampled variable \textbf{z} from a standard Gaussian distribution $\GaussianDistribution$. Thus, a latent representation $\tilde\xb_t=\generator(\xb_s, \modez)$ will generated by generator. Then discriminator will try to distinguish between $\tilde\xb_t$ and $\xb_t=E_{\text{AE}}(\Xb^t)$. The mode encoder $\Ez$ which is the encoder part of VAE will encode the synthetic output $\tilde\Xb^t$, which is decoded from the latent representation $\tilde\Xb^t=\Dcomp(\tilde\xb_t)$, to reconstruct the conditional input $\zrecon = \Ez(\tilde\Xb^t)$. 
\subsection{Train generator with classification loss}
The job of discriminator $F$ is to only distinguish the latent representation from the source domain or the target domain, and there is no guarantee that the synthetic output will be the same class as input object, though we use a reconstruction loss during training to make sure the output point set will be close to the input point set in Earth Mover’s Distance (EMD). So we add a discriminator to utilize label information from the source dataset fully. The classifier $\Cb$ will predict the class of output point set and compare it with the ground truth label in the source dataset. The loss will be backward to the generator to encourage it generates synthetic objects in the same class as input objects from the source dataset.

\subsection{Overall loss function and training }
To optimize GAN's output object quality, we set a min-max game between the generator, the discriminator, and the classifier. Given training examples of source domain object $\Xb^s$, and Gaussian samples $\modez$, we seek to optimize the following training losses over the generator $\Gb$, the discriminator $\Fb$, the encoder $\Eb_z$ and the classifier $\Cb$:

 \textbf{Adversarial loss.} For trianing of the generator and discriminator we add the an adversarial loss and we implement least square GAN~\cite{lsgan} for stabilizing the training. Hence, the adversarial losses will be minimized for the generator and the discriminator are defined as:
 \begin{align}
    \begin{split}
    \LossGAN_\discriminator & = \mathbb{E}_{\Xb^t \sim p(\Xb^t)}
    {[{\discriminator 
    {(  \Ecomp(\Xb^t) )}} -1 ]^2} \\ &
     +  \mathbb{E}_{\Xb^s \sim p(\Xb^s) } {[ {\discriminator{(
    \Ecomp ( \Xb^s ) }} 
   ]^2}  
   \end{split}
   \\
   \LossGAN_\generator & = 
   \mathbb{E}_{\Xb^s \sim p(\Xb^s), \modez \sim p(\modez) } {\left[
   \discriminator{\left(
   \generator( \Ecomp ( \Xb^s ), \modez ) \right)} -1
   \right]^2},
   \label{eqn:lsgan_loss}
\end{align}
where $\Xb^t \sim p(\Xb^t)$, $\Xb^s \sim p(\Xb^s)$ and $\modez \sim p(\modez)$ denotes samples drawn from the set of complete point sets, the set of partial point sets, and $\GaussianDistribution$.

\textbf{Reconstruction loss.}
To make the output object similar to the input object in shape, we add a reconstruction loss to encourage the generator to reconstruct the input so that the output object is more likely to be considered the same class as the input object. Here we use Earth  Mover’s  Distance (EMD) to measure the distance between the reconstructed object $\tilde\Xb$ and the input object $\Xb^s$.
\begin{equation}
    \LossRecon_\generator = \mathbb{E}_{\Xb^s \sim p(\Xb^s), \modez \sim p(\modez) } {\left[ d^{\text{EMD}}(\Xb^s, \Dcomp(\generator(\Ecomp(\Xb^s), \modez)) \right]} ),
    \label{eqn:partial_recon_loss}
\end{equation}

\textbf{Latent space reconstruction.}
A reconstruction loss on the $\modez$ latent space is also added to force $G$ to use the conditional mode vector $\modez$ in generate output object:
\begin{equation}
    \LossLatent_{{\generator, \Ez}} = \mathbb{E}_{\Xb^s \sim p(\Xb^s), \modez \sim p(\modez) } {\left[ \norm{\modez, \Ez(\Dcomp(\generator(\Ecomp(\Xb^s), \modez))) }_1 \right]},
    \label{eqn:partial_recon_loss}
\end{equation}

\textbf{Classification loss.}
To restrict the class of output object, we added the classification loss to the classifier and the generator.

\begin{align}
    &\LossCls_\classifier  = 
    \mathbb{E}_{\Xb^s \sim p(\Xb^s), \modez \sim p(\modez)}[-\sum_{k=1}^K l_k \log {(
   \generator( \Ecomp ( \Xb^s ), \modez ) )}]
   \\
   &\LossGAN_\generator  = 
   \mathbb{E}_{\Xb^s \sim p(\Xb^s), \modez \sim p(\modez)}[-\sum_{k=1}^K l_k \log {(
   \generator( \Ecomp ( \Xb^s ), \modez ) )}]
\end{align}

where $l_k$ is the $k$th label among all classes in source dataset
The full objective function for training the domain transfer network is described as:
\begin{equation}
    \argmin_{(\generator, \Ez,\classifier)} \argmax_{\discriminator} \LossGAN_\discriminator + \LossGAN_\generator + \alpha \LossRecon_\generator + \beta \LossLatent_{{\generator, \Ez}}+\gamma\LossCls_\generator,
    \label{eqn:obj_function}
\end{equation}
where $\alpha$, $\beta$ and $\gamma$ are importance weights for the reconstruction loss, the latent space reconstruction loss and classification loss respectively.

In choosing the importance weights, $\alpha$ controls how similar the shape of synthetic output object with the shape of the input object in 3D space, $\beta$ determines how close the synthetic output object with input object in latent space, and $\gamma$ influences how much probability the synthetic object will be considered as same class as input object under classifier.

\subsection{Network implementation}

In the experiments, each point cloud object is set to $1024$ points 
the VAE follows \cite{achlioptas2018learning,chen_pcl2pcl2020}: using  PointNet\cite{qi2017pointnet} as the encoder and a 3-layer MLP as the decoder. 
The autoencoder encodes a point set into a latent vector of fixed dimension $|\textbf{x}|=256$.
Similar to \cite{guo2020pct}, we use the 4-layer self attention layers with MLP as encoder and 3-layer MLP for both generator $\generator$ and discriminator $\discriminator$. Classifier is also using Pointnet structure.
To train the VAE, we use the Adam optimizer\cite{kingma2014adam} with an initial learning rate $0.0005$,   $\beta_1=0.9$ and train 2000 epochs with a batch size of 200. To train the autoencoder we use the Adam optimizer with an initial learning rate $0.0005$, $\beta_1=0.5$ and train for a maximum of 1000 epochs with a batch size of 32. The parameters of the pre-trained autoencoder and variational autoencoder are fixed during GAN training. To train the GAN, we use the Adam optimizer with an initial learning rate $0.0005$, $\beta_1=0.5$ and train for a maximum of 1000 epochs with a batch size of 50. The classifier used to train the GAN was pre-trained on the source dataset with the Adam optimizer in an initial learning rate $0.0001$ for 200 epochs.

\begin{table}[]
\begin{center}
\caption{Importance weights used in loss function  }\label{tab:parameter}
\begin{tabular}{|l|l|l|l|}
\hline
Dataset & $\alpha$ & $\beta$ & $\gamma$ \\ \hline
\textbf{M $\rightarrow$ S}     &0.05   &0.05   &0.01   \\ \hline
\textbf{M $\rightarrow$ S*}   &5  &5   &0.01   \\ \hline
\textbf{S $\rightarrow$ S*}   &10   &1   &0.01   \\ \hline
\textbf{S* $\rightarrow$ S}    &0.1   &0.1   &0.01   \\ \hline
\textbf{S $\rightarrow$ M}     &0.01   &0.01   &0.01   \\ \hline
\textbf{S* $\rightarrow$ M}    &0.05   &0.05   &0.01   \\ \hline
\end{tabular}
\end{center}
\end{table}

\begin{table*}[]
\begin{center}
\caption{Number of samples in proposed datasets. }\label{t2}
\label{tab:dataset}
\scalebox{0.8}{
\centering
\begin{tabular}{c cccccc cccccc}
\toprule
\multicolumn{2}{c}{Dataset}&\multicolumn{1}{c}{Bathtub}  &\multicolumn{1}{c}{Bed}&\multicolumn{1}{c}{Bookshelf}&\multicolumn{1}{c}{Cabinet}
&\multicolumn{1}{c}{Chair}  &\multicolumn{1}{c}{Lamp}&\multicolumn{1}{c}{Monitor}&\multicolumn{1}{c}{Plant}&\multicolumn{1}{c}{Sofa}&\multicolumn{1}{c}{Table}&\multicolumn{1}{c}{Total}\\
\midrule
\multirow{2}{*}{\textbf{M}}&Train&106 &515 &572 &200 &889&124 &465 &240&\multicolumn{1}{c}{680}&\multicolumn{1}{c}{392}  &\multicolumn{1}{c}{4, 183}\\
&Test&50 &100 &100  &86 &100&20 &100 &100 &\multicolumn{1}{c}{100}&\multicolumn{1}{c}{100}&\multicolumn{1}{c}{856}\\
\toprule
\multirow{2}{*}{\textbf{S}}&Train&\multicolumn{1}{c}{599} &\multicolumn{1}{c}{167} &\multicolumn{1}{c}{310}  &\multicolumn{1}{c}{1, 076} &\multicolumn{1}{c}{4, 612}&\multicolumn{1}{c}{1, 620} &\multicolumn{1}{c}{762}  &\multicolumn{1}{c}{158} &\multicolumn{1}{c}{2, 198}&\multicolumn{1}{c}{5, 876}  &\multicolumn{1}{c}{17, 378}\\
&Test&\multicolumn{1}{c}{85} &\multicolumn{1}{c}{23} &\multicolumn{1}{c}{50}  &\multicolumn{1}{c}{126} &\multicolumn{1}{c}{662}&\multicolumn{1}{c}{232} &\multicolumn{1}{c}{112}  &\multicolumn{1}{c}{30} &\multicolumn{1}{c}{330}&\multicolumn{1}{c}{842}  &\multicolumn{1}{c}{2, 492}\\
\toprule
\multirow{2}{*}{\textbf{S*}}&Train&\multicolumn{1}{c}{98} &\multicolumn{1}{c}{329} &\multicolumn{1}{c}{464}  &\multicolumn{1}{c}{650} &\multicolumn{1}{c}{2, 578}&\multicolumn{1}{c}{161} &\multicolumn{1}{c}{210}  &\multicolumn{1}{c}{88} &\multicolumn{1}{c}{495}&\multicolumn{1}{c}{1, 037}  &\multicolumn{1}{c}{6, 110}\\
&Test&\multicolumn{1}{c}{26} &\multicolumn{1}{c}{85} &\multicolumn{1}{c}{146}  &\multicolumn{1}{c}{149} &\multicolumn{1}{c}{801}&\multicolumn{1}{c}{41} &\multicolumn{1}{c}{61}  &\multicolumn{1}{c}{25} &\multicolumn{1}{c}{134}&\multicolumn{1}{c}{301}  &\multicolumn{1}{c}{1, 769}\\
\bottomrule 
\end{tabular}

}
\end{center}
\end{table*}

\section{Experiments}
\subsection{Datasets}
We verify our domain adaption model on three public point cloud datasets: shapenet \cite{chang2015shapenet}, scannet \cite{dai2017scannet} and modelnet \cite{wu20153d}. Following the same setting as PointDA-10  \cite{pointdan}, we choose ten common classes among three datasets. The results are shown in Table \ref{tab:dataset}, where \textbf{M}, \textbf{S} and \textbf{S*} represent subset of Modelnet, Shapenet and  Scannet respectively. We consider six types of adaptation scenarios which are \textbf{M $\rightarrow$ S}, \textbf{M $\rightarrow$ S*}, \textbf{S $\rightarrow$ M}, \textbf{S $\rightarrow$ S*}, \textbf{S* $\rightarrow$ M} and \textbf{S* $\rightarrow$ S}.
\begin{itemize}
\item[]
\textbf{ModelNet-10 (M):} ModelNet40 contains clean 3D CAD models of 40 categories. Those downloaded 3D CAD models from 3D Warehouse, and Yobi3D search engine indexing 261 CAD model websites. After getting the CAD model, they sample points on the surface to fully cover the object. To extract overlapped classes, we selected classes in ModelNet-10.
\item[]
\textbf{ShapeNet-10 (S):} ShapeNet contains 3D CAD models of 55 categories gathered from online repositories: Trimble 3D Warehouse and Yobi3D2. ShapeNet contains more samples and its objects have larger variance in structure compared with ModelNet. Uniform sampling is applied to collect the points of ShapeNet on surface. So  may lose some marginal points compared with ModelNet. Same as ModelNet-10 we collect 10 common classes from ShapeNet.
\item[]
\textbf{ScanNet-10 (S*):} Different with ModelNet and ShapeNet, ScanNet contains scanned and reconstructed real-world indoor scenes. We isolate 10 common classes instances contained in annotated bounding boxes for classification. The objects are scanned by RGB-D camera, they often lose some parts and get occluded by surroundings. ScanNet is a realistic domain but more challenging for DNN based model.

\end{itemize}

\subsection{Experiments Setup}
We choose the PointNet~\cite{qi2017pointnet} as the backbone of the classifier in evaluation. The learning rate is set to 0.0001 under the weight decay 0.0005 and $\alpha,\beta ,\gamma $ follow the Table \ref{tab:parameter}. 
All models have been trained for 200 epochs of batch size 64 in both source domain and synthetic datasets.

\begin{figure}
	\centering
	\includegraphics[width=0.95\columnwidth]{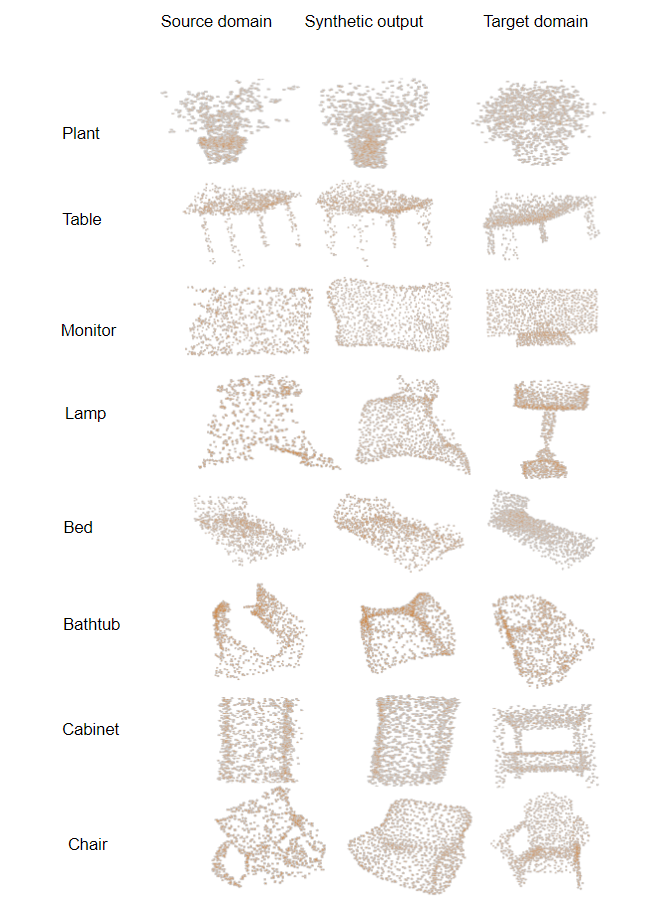}
	
	\caption{Visualization for source domain object, target domain object and synthetic object }
	\label{fig:demo result}
	\vspace{-0.3cm}
\end{figure}

\begin{table}[h]
\begin{center}
\caption{Quantitative classification results (\%) on PointDA-10 Dataset\cite{pointdan}. }\label{result}
\scalebox{0.7}{
\begin{threeparttable}
 \centering
  \begin{tabular}{ccccc ccccc cc}
\toprule 
\multicolumn{1}{c}{}&\multicolumn{1}{c}{M$\rightarrow$S}  &\multicolumn{1}{c}{M$\rightarrow$S*}&\multicolumn{1}{c}{S$\rightarrow$M}&\multicolumn{1}{c}{S$\rightarrow$S*}&\multicolumn{1}{c}{S*$\rightarrow$M}&\multicolumn{1}{c}{S*$\rightarrow$S}&\multicolumn{1}{c}{Avg}\\
\toprule
\multicolumn{1}{c}{w/o Adapt}  &\multicolumn{1}{c}{42.5} &\multicolumn{1}{c}{22.3} &\multicolumn{1}{c}{39.9}  &\multicolumn{1}{c}{23.5} &\multicolumn{1}{c}{34.2} &\multicolumn{1}{c}{46.9}  &\multicolumn{1}{c}{34.9}\\
\toprule  
\multicolumn{1}{c}{MMD\cite{long2013transfer}}  &\multicolumn{1}{c}{57.5} &\multicolumn{1}{c}{27.9} &\multicolumn{1}{c}{40.7}  &\multicolumn{1}{c}{26.7} &\multicolumn{1}{c}{47.3} &\multicolumn{1}{c}{54.8}  &\multicolumn{1}{c}{42.5}\\
\multicolumn{1}{c}{DANN\cite{ganin2014unsupervised}}   &\multicolumn{1}{c}{58.7} &\multicolumn{1}{c}{29.4} &\multicolumn{1}{c}{\textbf{42.3}}  &\multicolumn{1}{c}{30.5} &\multicolumn{1}{c}{48.1} &\multicolumn{1}{c}{56.7}  &\multicolumn{1}{c}{44.2}\\
\multicolumn{1}{c}{ADDA\cite{tzeng2017adversarial}}   &\multicolumn{1}{c}{61.0} &\multicolumn{1}{c}{30.5} &\multicolumn{1}{c}{40.4}  &\multicolumn{1}{c}{29.3} &\multicolumn{1}{c}{48.9} &\multicolumn{1}{c}{51.1}  &\multicolumn{1}{c}{43.5}\\
\multicolumn{1}{c}{MCD\cite{saito2018maximum}}    &\multicolumn{1}{c}{62.0} &\multicolumn{1}{c}{31.0} &\multicolumn{1}{c}{41.4}  &\multicolumn{1}{c}{31.3} &\multicolumn{1}{c}{46.8} &\multicolumn{1}{c}{59.3}  &\multicolumn{1}{c}{45.3}\\
\multicolumn{1}{c}{PointDAN\cite{pointdan}}   &\multicolumn{1}{c}{{62.5}} &\multicolumn{1}{c}{{31.2}} &\multicolumn{1}{c}{{41.5}}  &\multicolumn{1}{c}{{31.5}} &\multicolumn{1}{c}{{46.9}} &\multicolumn{1}{c}{{59.3}}  &\multicolumn{1}{c}{{45.5}}\\
\toprule
\multicolumn{1}{c}{Ours}  &\multicolumn{1}{c}{\textbf{62.8}} &\multicolumn{1}{c}{\textbf{36.5}} &\multicolumn{1}{c}{{41.9}} &\multicolumn{1}{c}{\textbf{31.6}} &\multicolumn{1}{c}{\textbf{50.4}} &\multicolumn{1}{c}{\textbf{65.7}}  &\multicolumn{1}{c}{\textbf{48.1}}\\

\toprule
\multicolumn{1}{c}{Supervised}  &\multicolumn{1}{c}{90.5} &\multicolumn{1}{c}{53.2} &\multicolumn{1}{c}{86.2}  &\multicolumn{1}{c}{53.2} &\multicolumn{1}{c}{86.2} &\multicolumn{1}{c}{90.5}  &\multicolumn{1}{c}{76.6}\\
\bottomrule  
\end{tabular}
\begin{tablenotes}

\item  \small \textbf{M} means ModelNet and \textbf{S} denotes ShapeNet while \textbf{S*} represents ScanNet.  
\end{tablenotes}
\end{threeparttable}
}
\end{center}
\end{table}

\textbf{Baselines:} In our experiments, we evaluate the performance when the model is trained only by source training samples (\textbf{w/o Adapt}). We also compare our model with five general-purpose UDA methods including: Maximum Mean Discrepancy (\textbf{MMD})~\cite{long2013transfer}, Adversarial Discriminative Domain Adaptation (\textbf{ADDA})~\cite{tzeng2017adversarial}, Domain Adversarial Neural Network (\textbf{DANN})~\cite{ganin2014unsupervised},  Maximum Classifier Discrepancy (\textbf{MCD})~\cite{saito2018maximum} and \textbf{PointDAN} \cite{pointdan} in the same training policy. For further performance comparison, we also evaluated the performance of an ablated version of our model that removes the classifier and only kept the generator and discriminator. Finally, we show the performance of a full supervised method (\textbf{Supervised}). As expected, supervised performs better than all unsupervised methods.

As shown from table \ref{result}, our approach can improve over all five existing UDA models in all six scenarios. In particular, for \textbf{S* $\rightarrow$ S}, our approach achieves 65.7\% predicting accuracy, which represents about 10.8\% improvement over the best result from  the existing approaches. For \textbf{S*$\rightarrow$M}, our model also has over 5.7\% improvement. Note that, among the three different datasets, Scannet (\textbf{S*}) is the 
most challenging dataset because Scannet's point cloud objects are scanned from the real world, while the other two datasets are generated from 3D polygonal models. This implies that the difference between \textbf{S*} and \textbf{S}, and between \textbf{S*} and \textbf{M} are more significant from the observation and our approach is better at generating synthetic objects by minimizing the gap in point distributions between the source and target datasets.

\begin{figure}[ht]
	\centering
	\includegraphics[width=1.0\columnwidth]{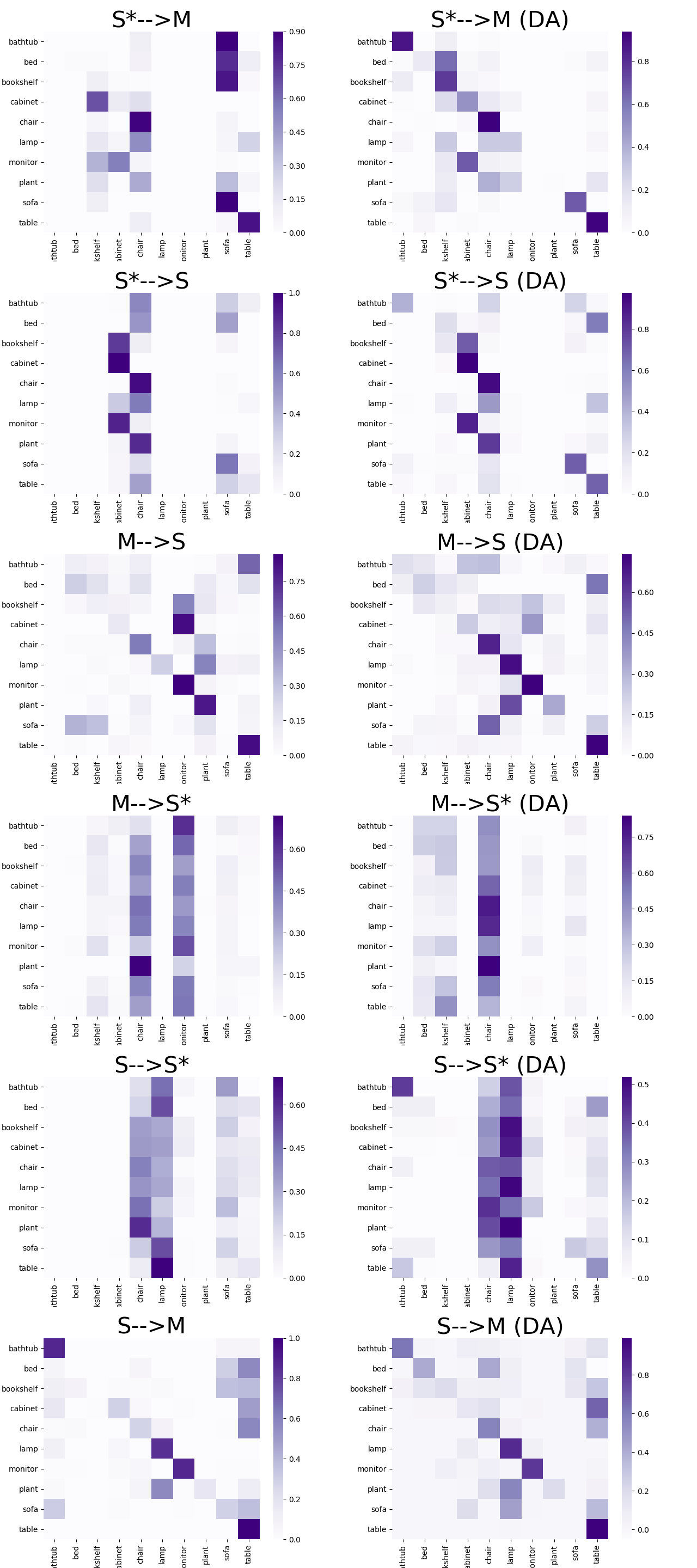}
	
	\caption{Confusion matrix }
	\label{fig:Confusion}
	\vspace{-0.3cm}
\end{figure}

To better understand how does our DA method improve the classifier over multi-class classification task. Confusion matrix over ten classes is shown in Fig.\ref{fig:Confusion}. Since each class has a different amount of objects, we normalize the amount to ratio. The higher value in the confusion matrix, the darker color it is. In Scannet(S*) to Modelnet(M) scenario, we can see a large amount of bathtub, bed, and bookshelf objects misclassified as a sofa. After domain adaption by our method, most of them are correctly classified. What's more, in  Modelnet(M) to Scannet(S*) scenario, which is a very challenging scenario. From table \ref{result} we find it has only 22.3\% accuracy without domain adaption, most of the objects are misclassified as chair or monitor. After applying our method to this scenario, the amount of objects that are misclassified as a monitor is significantly decreased, and we have more correctly classified objects in the bathtub and bed.

\begin{table}[h]
\begin{center}
\caption{Ablation analysis }\label{ablation}
\scalebox{0.6}{
\begin{threeparttable}
 \centering
  \begin{tabular}{ccccc ccccc cc}
\toprule 
\multicolumn{1}{c}{}&\multicolumn{1}{c}{\textbf{AE}} &\multicolumn{1}{c}{\textbf{L}}&\multicolumn{1}{c}{\textbf{C}} &\multicolumn{1}{c}{M$\rightarrow$S}  &\multicolumn{1}{c}{M$\rightarrow$S*}&\multicolumn{1}{c}{S$\rightarrow$M}&\multicolumn{1}{c}{S$\rightarrow$S*}&\multicolumn{1}{c}{S*$\rightarrow$M}&\multicolumn{1}{c}{S*$\rightarrow$S}&\multicolumn{1}{c}{Avg}\\
\toprule
\multicolumn{1}{c}{w/o Adapt} &{}  &\multicolumn{1}{c}{} &\multicolumn{1}{c}{}  &\multicolumn{1}{c}{42.5} &\multicolumn{1}{c}{22.3} &\multicolumn{1}{c}{39.9}  &\multicolumn{1}{c}{23.5} &\multicolumn{1}{c}{34.2} &\multicolumn{1}{c}{46.9}  &\multicolumn{1}{c}{34.9}\\
\toprule  
\multicolumn{1}{c}{only AE} &{$\surd$}  &\multicolumn{1}{c}{} &\multicolumn{1}{c}{}  &\multicolumn{1}{c}{59.5} &\multicolumn{1}{c}{33.5} &\multicolumn{1}{c}{34.2}  &\multicolumn{1}{c}{16.1} &\multicolumn{1}{c}{43.3} &\multicolumn{1}{c}{55.4}  &\multicolumn{1}{c}{40.3}\\
\multicolumn{1}{c}{AE+L} &{$\surd$}  &\multicolumn{1}{c}{$\surd$} &\multicolumn{1}{c}{}  &\multicolumn{1}{c}{62.6} &\multicolumn{1}{c}{34.1} &\multicolumn{1}{c}{40.4}  &\multicolumn{1}{c}{29.1} &\multicolumn{1}{c}{49.6} &\multicolumn{1}{c}{64.3}  &\multicolumn{1}{c}{46.7}\\
\multicolumn{1}{c}{GFA} &{$\surd$}  &\multicolumn{1}{c}{$\surd$} &\multicolumn{1}{c}{$\surd$}   &\multicolumn{1}{c}{\textbf{62.8}} &\multicolumn{1}{c}{\textbf{36.5}} &\multicolumn{1}{c}{\textbf{41.9}} &\multicolumn{1}{c}{\textbf{31.6}} &\multicolumn{1}{c}{\textbf{50.4}} &\multicolumn{1}{c}{\textbf{65.7}}  &\multicolumn{1}{c}{\textbf{48.1}}\\

\toprule
\multicolumn{1}{c}{Supervised} &{}  &\multicolumn{1}{c}{}  &\multicolumn{1}{c}{}  &\multicolumn{1}{c}{90.5} &\multicolumn{1}{c}{53.2} &\multicolumn{1}{c}{86.2}  &\multicolumn{1}{c}{53.2} &\multicolumn{1}{c}{86.2} &\multicolumn{1}{c}{90.5}  &\multicolumn{1}{c}{76.6}\\
\bottomrule  
\end{tabular}
\begin{tablenotes}

\item  \small \textbf{AE} means use autoencoder with reconstruction loss in model ,\textbf{L} denotes latent space reconstruction with VAE , \textbf{C} represents the additional discriminator, a classifier.  
\end{tablenotes}
\end{threeparttable}
}
\end{center}
\end{table}

\textbf{Ablation Study Setup:}
To study the effect of latent reconstruction module and additional discriminator, we first construct a model that only keeps encoder/decoder and generator. From table \ref{ablation} we can see this ablated model could only slightly improve the classifier over the none-adaption case. After applying the latent space reconstruction module to this ablated model, the classifier's performance significantly increases in all six scenarios due to the high fidelity, quality synthetic output. Even the ablated model can outperform the existing models most of the time, but adding classifier \textbf{C} to our model as an additional discriminator will further improve the result.

\begin{figure}[H]
	\centering
	\includegraphics[width=1.0\columnwidth]{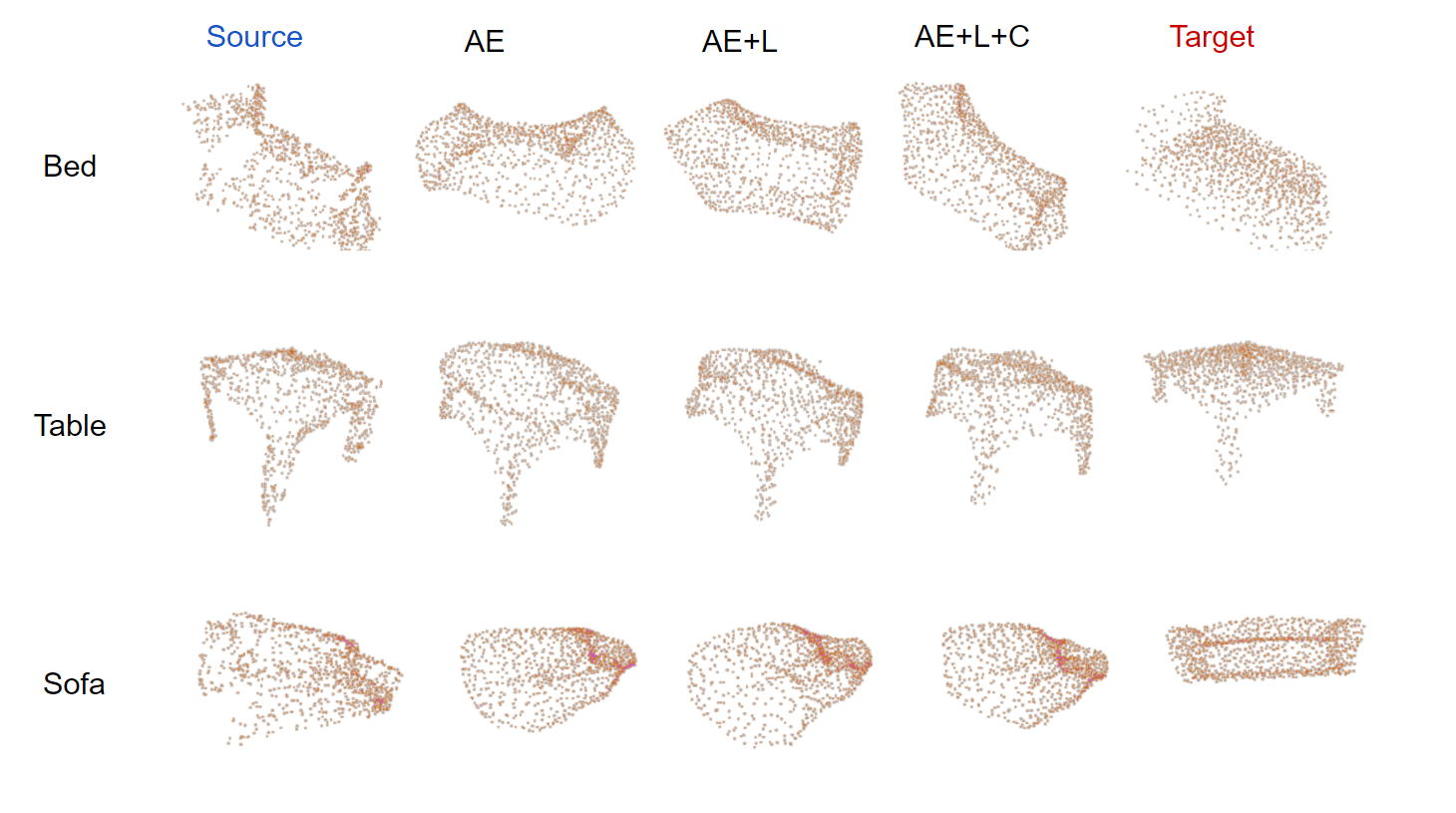}
	
	\caption{visualized result of different ablated models }
	\label{fig:ab_example}
	\vspace{-0.3cm}
\end{figure}

Fig.\ref{fig:ab_example} shows the visualized result of synthetic output generated by each ablated model from the same object in the source domain. We can find the model that only uses Autoencoder(AE) with reconstruction loss, can only roughly reconstruct the shape of the object. Nevertheless, after applying latent reconstruction loss with Variational Autoencoders(VAE), the quality of synthetic objects is significantly increased. In the last, using the classifier as an additional discriminator provides the synthetic objects more fidelity because it will encourage the generator to ingratiate the classifier so that the synthetic is more likely to be considered as the same class as the source domain object.

\section{Conclusion}
We have proposed a novel approach to unsupervised domain adaptation in the 3D classification task in this work. The basic idea is to transfer training data into the target domain rather than selecting domain invariant feature or implementing feature alignment. In our approach, a Generative Adversarial Network (GAN) is constructed for domain transfer in 3D point-clouds to perform unsupervised domain adaptation in 3D classification. Our model generates synthetic objects from the source domain objects such that the output will have the same shape and label as the source domain objects but are constructed according to the pattern of the target domain. In this way, classifier trained using the synthetic dataset will perform better in the target domain dataset. To encourage the model to generate an object in the same class as input, we designed a GAN-based framework that takes the classifier as an addition discriminator.

Using extensive experiments involving three datasets and comparing them with five existing DA methods, we have demonstrated our approach's superiority over the state-of-the-art domain adaptation methods.

{\small
\bibliographystyle{ieee_fullname}
\bibliography{egpaper_final}

\begin{thebibliography}{10}\itemsep=-1pt

\bibitem{achlioptas2018learning}
Panos Achlioptas, Olga Diamanti, Ioannis Mitliagkas, and Leonidas Guibas.
\newblock Learning representations and generative models for 3d point clouds.
\newblock pages 40--49, 2018.

\bibitem{Atzmon2018point}
Matan Atzmon, Haggai Maron, and Yaron Lipman.
\newblock Point convolutional neural networks by extension operators.
\newblock {\em {ACM} Transactions on Graphics}, 37(4):71:1--71:12, 2018.

\bibitem{borgwardt2006integrating}
Karsten~M Borgwardt, Arthur Gretton, Malte~J Rasch, Hans-Peter Kriegel,
  Bernhard Sch{\"o}lkopf, and Alex~J Smola.
\newblock Integrating structured biological data by kernel maximum mean
  discrepancy.
\newblock {\em Bioinformatics}, 22(14):e49--e57, 2006.

\bibitem{bousmalis2017unsupervised}
Konstantinos Bousmalis, Nathan Silberman, David Dohan, Dumitru Erhan, and Dilip
  Krishnan.
\newblock Unsupervised pixel-level domain adaptation with generative
  adversarial networks.
\newblock In {\em Proceedings of the IEEE conference on computer vision and
  pattern recognition}, pages 3722--3731, 2017.

\bibitem{chang2015shapenet}
Angel~X Chang, Thomas Funkhouser, Leonidas Guibas, Pat Hanrahan, Qixing Huang,
  Zimo Li, Silvio Savarese, Manolis Savva, Shuran Song, Hao Su, et~al.
\newblock Shape{N}et: An information-rich {3D} model repository.
\newblock {\em arXiv preprint arXiv:1512.03012}, 2015.

\bibitem{chen_pcl2pcl2020}
Xuelin Chen, Baoquan Chen, and Niloy~J. Mitra.
\newblock Unpaired point cloud completion on real scans using adversarial
  training.
\newblock 2020.

\bibitem{chen2020}
Mitra~NJ Chen~X, Chen~B.
\newblock Unpaired point cloud completion on real scans using adversarial
  training.
\newblock {\em International Conference on Learning Representations (ICLR)},
  2020.

\bibitem{dai2017scannet}
Angela Dai, Angel~X Chang, Manolis Savva, Maciej Halber, Thomas Funkhouser, and
  Matthias Niener.
\newblock Scan{N}et: Richly annotated {3D} reconstructions of indoor scenes.
\newblock In {\em Proceedings of the IEEE Conference on Computer Vision and
  Pattern Recognition}, pages 5828--5839, 2017.

\bibitem{feng2019meshnet}
Yutong Feng, Yifan Feng, Haoxuan You, Xibin Zhao, and Yue Gao.
\newblock Mesh{N}et: mesh neural network for {3D} shape representation.
\newblock In {\em Proceedings of the AAAI Conference on Artificial
  Intelligence}, volume~33, pages 8279--8286, 2019.

\bibitem{fernando2013unsupervised}
Basura Fernando, Amaury Habrard, Marc Sebban, and Tinne Tuytelaars.
\newblock Unsupervised visual domain adaptation using subspace alignment.
\newblock In {\em Proceedings of the IEEE international conference on computer
  vision}, pages 2960--2967, 2013.

\bibitem{ganin2014unsupervised}
Yaroslav Ganin and Victor Lempitsky.
\newblock Unsupervised domain adaptation by backpropagation.
\newblock {\em arXiv preprint arXiv:1409.7495}, 2014.

\bibitem{ganin2016domain}
Yaroslav Ganin, Evgeniya Ustinova, Hana Ajakan, Pascal Germain, Hugo
  Larochelle, Fran{\c{c}}ois Laviolette, Mario Marchand, and Victor Lempitsky.
\newblock Domain-adversarial training of neural networks.
\newblock {\em The Journal of Machine Learning Research}, 17(1):2096--2030,
  2016.

\bibitem{gong2013connecting}
Boqing Gong, Kristen Grauman, and Fei Sha.
\newblock Connecting the dots with landmarks: Discriminatively learning
  domain-invariant features for unsupervised domain adaptation.
\newblock In {\em Proceedings of the International Conference on Machine
  Learning}, pages 222--230, 2013.

\bibitem{gong2017geodesic}
Boqing Gong, Kristen Grauman, and Fei Sha.
\newblock Geodesic flow kernel and landmarks: Kernel methods for unsupervised
  domain adaptation.
\newblock In {\em Domain Adaptation in Computer Vision Applications}, pages
  59--79. Springer, 2017.

\bibitem{gopalan2011domain}
Raghuraman Gopalan, Ruonan Li, and Rama Chellappa.
\newblock Domain adaptation for object recognition: An unsupervised approach.
\newblock In {\em Proceedings of the IEEE International Conference on Computer
  Vision}, pages 999--1006, 2011.

\bibitem{guo2020pct}
Meng-Hao Guo, Jun-Xiong Cai, Zheng-Ning Liu, Tai-Jiang Mu, Ralph~R. Martin, and
  Shi-Min Hu.
\newblock Pct: Point cloud transformer, 2020.

\bibitem{hermosilla2018monte}
Pedro Hermosilla, Tobias Ritschel, Pere{-}Pau V{\'{a}}zquez, {\`{A}}lvar
  Vinacua, and Timo Ropinski.
\newblock Monte carlo convolution for learning on non-uniformly sampled point
  clouds.
\newblock {\em {ACM} Transactions on Graphics}, 37(6):235:1--235:12, 2018.

\bibitem{hoffman2017cycada}
Judy Hoffman, Eric Tzeng, Taesung Park, Jun-Yan Zhu, Phillip Isola, Kate
  Saenko, Alexei~A Efros, and Trevor Darrell.
\newblock Cycada: Cycle-consistent adversarial domain adaptation.
\newblock {\em arXiv preprint arXiv:1711.03213}, 2017.

\bibitem{kingma2014adam}
Diederik~P Kingma and Jimmy Ba.
\newblock Adam: A method for stochastic optimization.
\newblock {\em arXiv preprint arXiv:1412.6980}, 2014.

\bibitem{liu2019generative}
Lanlan Liu, Michael Muelly, Jia Deng, Tomas Pfister, and Li-Jia Li.
\newblock Generative modeling for small-data object detection, 2019.

\bibitem{liu2019pointvoxel}
Zhijian Liu, Haotian Tang, Yujun Lin, and Song Han.
\newblock Point-voxel cnn for efficient 3d deep learning, 2019.

\bibitem{long2013transfer}
Mingsheng Long, Jianmin Wang, Guiguang Ding, Jiaguang Sun, and Philip~S Yu.
\newblock Transfer feature learning with joint distribution adaptation.
\newblock In {\em Proceedings of IEEE International Conference on Computer
  Vision}, 2013.

\bibitem{lsgan}
Xudong Mao, Qing Li, Haoran Xie, Raymond~YK Lau, Zhen Wang, and Stephen
  Paul~Smolley.
\newblock Least squares generative adversarial networks.
\newblock pages 2794--2802, 2017.

\bibitem{maturana2015voxnet}
Daniel Maturana and Sebastian Scherer.
\newblock Voxnet: A 3d convolutional neural network for real-time object
  recognition.
\newblock In {\em Intelligent Robots and Systems (IROS), 2015 IEEE/RSJ
  International Conference on}, pages 922--928. IEEE, 2015.

\bibitem{nguyen2017dual}
Tu~Dinh Nguyen, Trung Le, Hung Vu, and Dinh Phung.
\newblock Dual discriminator generative adversarial nets, 2017.

\bibitem{qi2017pointnet}
Charles~R Qi, Hao Su, Kaichun Mo, and Leonidas~J Guibas.
\newblock Pointnet: Deep learning on point sets for 3d classification and
  segmentation.
\newblock {\em Proc. Computer Vision and Pattern Recognition (CVPR), IEEE},
  1(2):4, 2017.

\bibitem{qi2017pointnet++}
Charles~Ruizhongtai Qi, Li Yi, Hao Su, and Leonidas~J Guibas.
\newblock Pointnet++: Deep hierarchical feature learning on point sets in a
  metric space.
\newblock In {\em Advances in Neural Information Processing Systems}, pages
  5099--5108, 2017.

\bibitem{pointdan}
Can Qin, Haoxuan You, Lichen Wang, C.-C.~Jay Kuo, and Yun Fu.
\newblock Pointdan: A multi-scale 3d domain adaption network for point cloud
  representation.
\newblock In H. Wallach, H. Larochelle, A. Beygelzimer, F. d\textquotesingle
  Alch\'{e}-Buc, E. Fox, and R. Garnett, editors, {\em Advances in Neural
  Information Processing Systems 32}, pages 7190--7201. Curran Associates,
  Inc., 2019.

\bibitem{rist2019cross}
Christoph~B Rist, Markus Enzweiler, and Dariu~M Gavrila.
\newblock Cross-sensor deep domain adaptation for lidar detection and
  segmentation.
\newblock In {\em 2019 IEEE Intelligent Vehicles Symposium (IV)}, pages
  1535--1542. IEEE, 2019.

\bibitem{saito2018maximum}
Kuniaki Saito, Kohei Watanabe, Yoshitaka Ushiku, and Tatsuya Harada.
\newblock Maximum classifier discrepancy for unsupervised domain adaptation.
\newblock In {\em Proceedings of the IEEE Conference on Computer Vision and
  Pattern Recognition}, pages 3723--3732, 2018.

\bibitem{saleh2019domain}
Khaled Saleh, Ahmed Abobakr, Mohammed Attia, Julie Iskander, Darius Nahavandi,
  and Mohammed Hossny.
\newblock Domain adaptation for vehicle detection from bird's eye view {LiDAR}
  point cloud data.
\newblock {\em arXiv preprint arXiv:1905.08955}, 2019.

\bibitem{shrivastava2017learning}
Ashish Shrivastava, Tomas Pfister, Oncel Tuzel, Joshua Susskind, Wenda Wang,
  and Russell Webb.
\newblock Learning from simulated and unsupervised images through adversarial
  training.
\newblock In {\em Proceedings of the IEEE conference on computer vision and
  pattern recognition}, pages 2107--2116, 2017.

\bibitem{su2015multi}
Hang Su, Subhransu Maji, Evangelos Kalogerakis, and Erik Learned-Miller.
\newblock Multi-view convolutional neural networks for 3d shape recognition.
\newblock In {\em Proceedings of the IEEE international conference on computer
  vision}, pages 945--953, 2015.

\bibitem{sugiyama2008direct}
Masashi Sugiyama, Shinichi Nakajima, Hisashi Kashima, Paul~V Buenau, and
  Motoaki Kawanabe.
\newblock Direct importance estimation with model selection and its application
  to covariate shift adaptation.
\newblock In {\em Proceedings of the Advances in Neural Information Processing
  Systems}, pages 1433--1440, 2008.

\bibitem{sun2016deep}
Baochen Sun and Kate Saenko.
\newblock Deep coral: Correlation alignment for deep domain adaptation.
\newblock In {\em Proceedings of the European Conference on Computer Vision},
  2016.

\bibitem{Tchapmi2017segcloud}
Lyne~P. Tchapmi, Christopher~B. Choy, Iro Armeni, JunYoung Gwak, and Silvio
  Savarese.
\newblock {SEGCloud}: Semantic segmentation of 3d point clouds.
\newblock In {\em International Conference on 3D Vision}, pages 537--547.
  {IEEE} Computer Society, 2017.

\bibitem{tzeng2017adversarial}
Eric Tzeng, Judy Hoffman, Kate Saenko, and Trevor Darrell.
\newblock Adversarial discriminative domain adaptation.
\newblock In {\em Proceedings of the IEEE conference on Computer Vision and
  Pattern Recognition}, 2017.

\bibitem{wu2019squeezesegv2}
Bichen Wu, Xuanyu Zhou, Sicheng Zhao, Xiangyu Yue, and Kurt Keutzer.
\newblock Squeeze{S}eg{V}2: Improved model structure and unsupervised domain
  adaptation for road-object segmentation from a lidar point cloud.
\newblock In {\em Proceedings of the International Conference on Robotics and
  Automation}, pages 4376--4382, 2019.

\bibitem{wu_2020_ECCV}
Rundi Wu, Xuelin Chen, Yixin Zhuang, and Baoquan Chen.
\newblock Multimodal shape completion via conditional generative adversarial
  networks.
\newblock In {\em The European Conference on Computer Vision (ECCV)}, August
  2020.

\bibitem{Wu2019pointconv}
Wenxuan Wu, Zhongang Qi, and Fuxin Li.
\newblock {PointConv}: Deep convolutional networks on 3d point clouds.
\newblock In {\em {IEEE/CVF} Conference on Computer Vision and Pattern
  Recognition}, pages 9621--9630, 2019.

\bibitem{wu20153d}
Zhirong Wu, Shuran Song, Aditya Khosla, Fisher Yu, Linguang Zhang, Xiaoou Tang,
  and Jianxiong Xiao.
\newblock 3d shapenets: A deep representation for volumetric shapes.
\newblock In {\em Proceedings of the IEEE conference on computer vision and
  pattern recognition}, pages 1912--1920, 2015.

\bibitem{yi2020complete}
Li Yi, Boqing Gong, and Thomas Funkhouser.
\newblock Complete and label: A domain adaptation approach to semantic
  segmentation of lidar pointclouds, 2020.

\bibitem{you2018pvnet}
Haoxuan You, Yifan Feng, Rongrong Ji, and Yue Gao.
\newblock {PVNet}: A joint convolutional network of point cloud and multi-view
  for 3d shape recognition.
\newblock In {\em Proceedings of the ACM Multimedia Conference on Multimedia
  Conference}, pages 1310--1318, 2018.

\bibitem{zhang2013domain}
Kun Zhang, Bernhard Sch{\"o}lkopf, Krikamol Muandet, and Zhikun Wang.
\newblock Domain adaptation under target and conditional shift.
\newblock In {\em Proceedings of the International Conference on Machine
  Learning}, pages 819--827, 2013.

\end{thebibliography}
}

\end{document}